\documentclass[sigconf,review=false]{acmart}

\usepackage{booktabs}
\usepackage{multirow}
\usepackage{array}
\usepackage{xcolor}
\usepackage{tikz}
\usetikzlibrary{positioning,arrows.meta}

\copyrightyear{2026}
\acmYear{2026}
\setcopyright{cc}
\setcctype{by}
\acmConference[SIGSPATIAL '26]{The 34th ACM International Conference on Advances in Geographic Information Systems}{November 03--06, 2026}{Riverside, CA, USA}
\acmBooktitle{The 34th ACM International Conference on Advances in Geographic Information Systems (SIGSPATIAL '26), November 03--06, 2026, Riverside, CA, USA}
\acmDOI{10.1145/3841645.3843355}
\acmISBN{979-8-4007-2950-8/2026/11}

\begin{document}

\title[ML for Pedestrian Volume Estimation]{Estimating Pedestrian Volumes from GIS-Derived Built-Environment Features: A Machine Learning Framework}

\author{Bahareh Golchin}
\affiliation{%
  \institution{Department of Computer Science, \\Portland State University}
  \city{Portland}
  \state{Oregon}
  \country{USA}
}
\email{bgolchin@pdx.edu}

\author{Banafsheh Rekabdar}
\affiliation{%
  \institution{Department of Computer Science, \\Portland State University}
  \city{Portland}
  \state{Oregon}
  \country{USA}
}
\email{rekabdar@pdx.edu}

\author{Sirisha Kothuri}
\affiliation{%
  \institution{Dept.\ of Civil and Environmental Engineering, \\Portland State University}
  \city{Portland}
  \state{Oregon}
  \country{USA}
}
\email{skothuri@pdx.edu}

\author{Joseph Broach}
\affiliation{%
  \institution{Transportation Research and Education Center (TREC), \\Portland State University}
  \city{Portland}
  \state{Oregon}
  \country{USA}
}
\email{jbroach@pdx.edu}

\renewcommand{\shortauthors}{Golchin, Rekabdar, Kothuri, and Broach}
\begin{abstract}
Transportation agencies need pedestrian volume estimates across entire road networks to prioritize safety investments, yet manual counts are expensive and cover only a small share of intersections. We present a machine learning pipeline that predicts $2$-hour PM peak pedestrian volume at $101$ urban intersections in Portland, Oregon, from built-environment, land-use, and street-network features drawn from open GIS data. Starting from the Negative Binomial GLM used in practice, we add feature selection, count-aware gradient boosting, and repeated cross-validation, selecting one configuration by a combined rank over RMSE, MAPE, and SMAPE across four cross-validation strategies. The winner, a histogram-based gradient boosting model with Poisson loss and $L_1$ Lasso feature selection, reduces cross-validated RMSE by $12\%$ over the GLM baseline ($89.8 \rightarrow 78.7$) and holdout RMSE by $19\%$ ($108.0 \rightarrow 87.9$). Code is released on GitHub\footnote{\url{https://github.com/baharehgl/Pedestrian-Count-Estimations}}.
\end{abstract}




\maketitle

\section{Introduction}
Pedestrian volumes are among the least systematically measured quantities in urban transportation systems. Without reliable spatial estimates of walking activity, agencies cannot prioritize crossing improvements, evaluate the pedestrian impact of land-use changes, or quantify exposure when modeling crash risk~\cite{schneider2009pilot}. Manual counting is expensive and spatially sparse and automated counters are costly to maintain, so most jurisdictions hold counts at a small fraction of locations of interest. Direct demand models built on open GIS-derived variables are an attractive complement: a model that generalizes from counted sites can produce estimates anywhere in the network at marginal cost.

The standard workhorse is a Generalized Linear Model (GLM) with Negative Binomial errors, suited to overdispersed counts~\cite{schneider2009pilot}. It is interpretable but imposes strong functional-form assumptions and cannot represent nonlinear interactions among built-environment variables. Machine learning methods that can~\cite{breiman2001rf,friedman2001gbm,wang2022machine} struggle in the small-sample regime where most pedestrian count datasets sit; as in other low-label settings, the binding constraint is the validation protocol rather than model capacity~\cite{golchin2026llm,golchin2025dynamic}. Our dataset has $n{=}101$ sites and $79$ candidate predictors ($92$ after one-hot encoding), a regime in which using all features produces severe overfitting (Section~\ref{sec:results}). We contribute an end-to-end pipeline that takes this regime seriously, combining feature selection~\cite{tibshirani1996lasso}, repeated cross-validation (CV), and tuning, with a combined-rank procedure that avoids cherry-picking a metric. We benchmark seven models, four selectors, four CV strategies, and five feature counts across roughly $70{,}000$ fits.

\section{Data and Study Area}\label{sec:data}

\paragraph{Data Sources and Pedestrian Counts.}
In this study, Strava and static network and sociodemographic data sources were used to estimate 2-hr peak hour observed pedestrian intersection counts. Strava Metro provided GPS-traced pedestrian (walk and run) activity mapped to an OpenStreetMap (OSM) street network, though privacy masking and data sparsity at finer time scales limited its usefulness to annual totals. StreetLight, which processes location-based smartphone data into mode-specific trip estimates, had previously offered link-level bicycle and pedestrian data, but changes to smartphone privacy protocols in early 2022 disrupted that capability; a newer census tract-level product was not accessible in time for this study. Manual pedestrian counts were extracted from 192 PDF files provided by the City of Portland Bureau of Transportation, focusing on PM peak hour volumes (4--6 PM) at $101$ locations, though these were not randomly distributed across the city and were conducted predominantly in spring and summer.
Figure~\ref{fig:count_locations} maps the $101$ count locations across the City of Portland, with marker size scaled to observed $2$-hour PM peak entering volume.

\paragraph{Intersection Geocoding and Buffer Definition.}
Count locations were geocoded using QGIS, with roughly $10\%$ requiring manual placement due to geocoding failures or spatial misalignment. Rather than attempting to precisely identify intersection nodes, which proved inconsistent across locations, each intersection was defined as a $100$-foot radius buffer around the inferred count point. This approach balanced automation and replicability while capturing most of the relevant legs and crossing features of each location.

\paragraph{Feature Construction.}
Strava data were aligned to these intersection buffers using a distance-normalization method to avoid double-counting trips on the dense, multi-segment pedestrian network. By summing distance traveled within each buffer and dividing by an assumed $200$-foot traversal length, the team estimated total intersection-level Strava trips traversing the location. Background Strava activity was also measured within larger buffers ($1/8$, $1/4$, and $1/2$ mile) as potential model covariates. Static built environment and sociodemographic variables were drawn from OSM, the American Community Survey, and Census Longitudinal Employer-Household Dynamics (LEHD) data, and were measured both at the immediate intersection-area level and within the same series of distance buffers, with area- or household-weighted aggregation as appropriate. The OSMNX tool~\cite{boeing2017osmnx} was used to extract intersection nodes from the raw OSM data. The OSMNX is a Python package that converts OSM data into routable networks.

\begin{figure}[!t]
\centering
\includegraphics[width=0.70\linewidth]{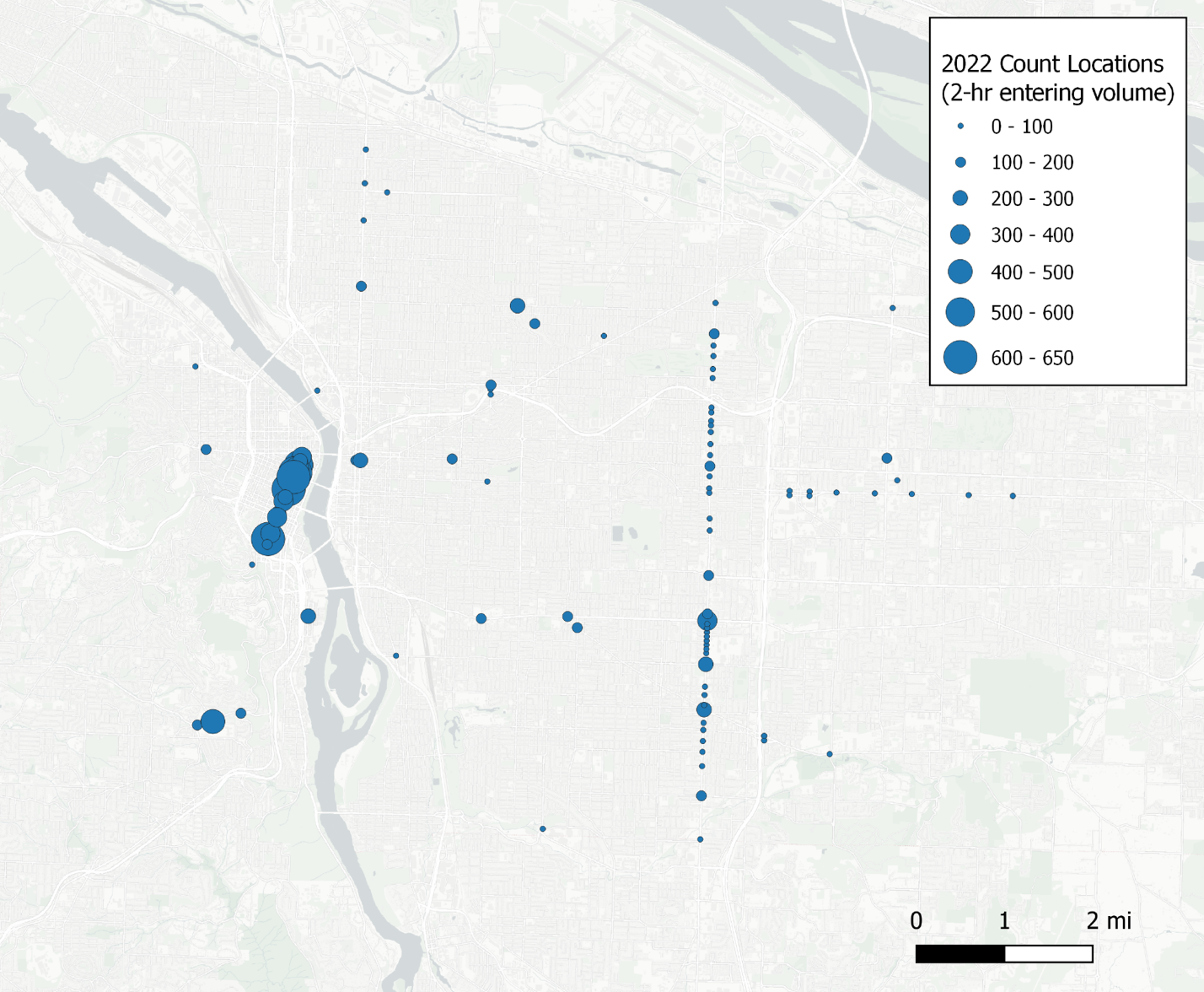}
\caption{The $101$ count locations, sized by observed $2$-hour PM peak entering volume. Coverage concentrates downtown and along project corridors rather than being randomly sampled.}
\Description{Map of Portland, Oregon with 101 count locations shown as circles scaled by 2-hour PM peak entering volume from 0 to 650 pedestrians; the largest markers cluster in the downtown core and along major north-south corridors.}
\label{fig:map}
\end{figure}

\section{Methodology}\label{sec:method}
Let $\mathbf{x}_i \in \mathbb{R}^p$ denote the GIS-derived feature vector at site $i$ and $y_i \in \mathbb{Z}_{\geq 0}$ the observed 2-hour PM peak count. We seek $\hat{f}: \mathbb{R}^p \rightarrow \mathbb{R}_{\geq 0}$ that generalizes to unseen sites. The loss landscape is dominated by a few high-volume urban intersections that drive RMSE, while a long tail of low-volume sites drives percentage errors.

\subsection{Baseline and candidate models}
Following~\cite{schneider2009pilot}, the baseline is a Negative Binomial regression,
\begin{equation}
y_i \mid \mathbf{x}_i \sim \mathrm{NB}(\mu_i, \theta), \qquad \log \mu_i = \beta_0 + \mathbf{x}_i^{\top}\boldsymbol{\beta},
\end{equation}
where $\theta$ governs overdispersion and $(\boldsymbol{\beta},\theta)$ are estimated by maximum likelihood. The published R-code analysis evaluated three hand-curated feature sets of $3$, $8$, and $10$ variables; the richest is the primary baseline throughout.

We evaluate seven models spanning the families relevant to count prediction on small tabular data. Random Forest~\cite{breiman2001rf} is a bagging baseline, robust to outliers and low-variance. HistGB\_Poisson is histogram-based gradient boosting with Poisson loss: its log-link enforces non-negative predictions and shares the distributional assumption of the GLM, while the tree ensemble adds nonlinearity, making it the closest counterpart to the baseline. HistGB with squared loss isolates the effect of the loss function. Extra Trees~\cite{geurts2006extremely} reduces variance at small $n$; Huber-loss boosting is robust to outlying high-volume sites; a regularized Poisson GLM is an interpretable reference; and bagged HistGB tests variance reduction on a strong learner.

\subsection{Feature selection}
With at most $81$ training sites and $92$ encoded features, selection is mandatory. We compare four methods. The $L_1$ Lasso~\cite{tibshirani1996lasso} adds an $\ell_1$ penalty,
\begin{equation}
\hat{\boldsymbol{\beta}}^{\mathrm{lasso}} = \arg\min_{\boldsymbol{\beta}} \tfrac{1}{2n}\|\mathbf{y}-\mathbf{X}\boldsymbol{\beta}\|_2^2 + \lambda\|\boldsymbol{\beta}\|_1,
\end{equation}
driving many coefficients to exactly zero and producing a sparse set whose non-zero entries are interpretable on the log-link scale. Mutual information measures any (not necessarily linear) dependence between feature $X_j$ and target $Y$, $I(X_j;Y)=\sum_{x,y} p(x,y)\log\frac{p(x,y)}{p(x)p(y)}$, estimated by $k$-nearest-neighbor density estimation. Random Forest importance ranks features by mean decrease in impurity across the ensemble, capturing interaction-driven contributions. F-regression is a univariate F-test included as a sanity-check lower bound. All selectors are fit on training folds only, so the holdout never informs selection.

\subsection{Validation and tuning}
With $n{=}101$, a single $20$-site holdout is noisy: a few high-volume sites shift RMSE by $\pm 20$, making comparison unreliable. Cross-validation instead rotates every site through the test set. We evaluate four strategies: $5$-fold ($\sim$20 test samples per fold, high variance); $10$-fold ($\sim$10 per fold); repeated $5{\times}10$; and repeated $3{\times}10$, averaging over $30$ folds that cover every site three times. Repeated $3{\times}10$ gave the smallest standard deviation of RMSE across the $32$ cells of the design and is the estimator behind all headline results. All preprocessing (imputation, scaling, one-hot encoding) is fit on the training folds of each split to prevent leakage. For each model we sample $40$ hyperparameter combinations via \texttt{RandomizedSearchCV}~\cite{pedregosa2011scikit} with inner cross-validation, over ranges chosen for overfitting control on small tabular data: learning rate $\{0.01,0.05,0.1\}$, boosting rounds $\{300,500,800\}$, depth $\{5,7,10\}$, minimum leaf size $\{10,20\}$, $\ell_2$ penalty $\{0,1\}$, and, for the bagging and forest models, $300$--$800$ estimators.

\subsection{Evaluation metrics}
Three metrics are reported in parallel, each capturing a different property of the residual distribution: $\text{RMSE}=\sqrt{\frac{1}{n}\sum_i (y_i-\hat{y}_i)^2}$, $\text{MAPE}=\frac{100}{n}\sum_i \frac{|y_i-\hat{y}_i|}{\max(|y_i|,\epsilon)}$, and $\text{SMAPE}=\frac{100}{n}\sum_i \frac{2|y_i-\hat{y}_i|}{|y_i|+|\hat{y}_i|+\epsilon}$. RMSE is dominated by high-volume sites, MAPE by low-volume sites, and SMAPE is bounded and stable across the range. McFadden's pseudo-$R^2$ is reported on training folds as an in-sample diagnostic only; it is not comparable across model families.

\section{Experiments and Results}\label{sec:results}
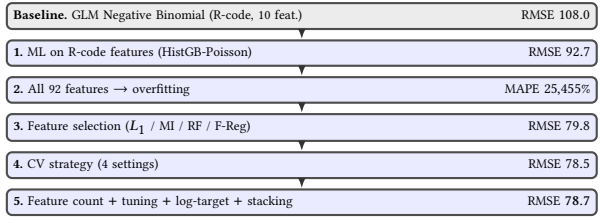
\begin{figure}[!t]
\centering
\begin{tikzpicture}[
  node distance=3.5pt,
  every node/.style={font=\tiny},
  stage/.style={
     rectangle, rounded corners=2pt, draw=black!70, thick,
     text width=0.90\linewidth, inner sep=2.5pt,
     align=left, fill=blue!8
  },
  base/.style={stage, fill=gray!15},
  arrow/.style={-{Latex[length=1.4mm]}, thick, black!80}
]
\node[base] (p0) {\textbf{Baseline.} GLM Negative Binomial (R-code, 10 feat.) \hfill RMSE $108.0$};
\node[stage, below=of p0] (p1) {\textbf{1.} ML on R-code features (HistGB-Poisson) \hfill RMSE $92.7$};
\node[stage, below=of p1] (p2) {\textbf{2.} All 92 features $\rightarrow$ overfitting \hfill MAPE $25{,}455\%$};
\node[stage, below=of p2] (p3) {\textbf{3.} Feature selection ($L_1$ / MI / RF / F-Reg) \hfill RMSE $79.8$};
\node[stage, below=of p3] (p4) {\textbf{4.} CV strategy (4 settings) \hfill RMSE $78.5$};
\node[stage, below=of p4] (p5) {\textbf{5.} Feature count $+$ tuning $+$ log-target $+$ stacking \hfill RMSE $\mathbf{78.7}$};
\foreach \a/\b in {p0/p1, p1/p2, p2/p3, p3/p4, p4/p5}
  \draw[arrow] (\a) -- (\b);
\end{tikzpicture}
\caption{Pipeline progression. The gray box is the published GLM baseline; each numbered phase removes one source of error. Phase 2 reports MAPE because GLM RMSE diverges. Baseline and Phases 1--2 are holdout; Phases 3--5 are cross-validated.}
\Description{A vertical flow diagram of six stacked boxes. A gray baseline box at the top shows RMSE 108.0, followed by five blue phase boxes ending at cross-validated RMSE 78.7.}
\label{fig:pipeline}
\end{figure}

Figure~\ref{fig:pipeline} summarizes our pipeline.
Table~\ref{tab:main} traces the pipeline, Table~\ref{tab:ablate} compares feature selectors, and Table~\ref{tab:tuning} reports the tuning sweep.

\begin{table}[!t]
\caption{Pipeline progression. All configurations evaluated on the identical $81/20$ split and under repeated $3{\times}10$ CV.}
\label{tab:main}
\centering
\footnotesize
\setlength{\tabcolsep}{4pt}
\renewcommand{\arraystretch}{1.05}
\begin{tabular}{lcccc}
\toprule
& \multicolumn{2}{c}{\textbf{Holdout}} & \multicolumn{2}{c}{\textbf{Rep.\ $3{\times}10$ CV}} \\
\cmidrule(lr){2-3}\cmidrule(lr){4-5}
\textbf{Configuration} & RMSE & MAPE & RMSE & MAPE \\
\midrule
GLM-NegBin (R-code, 10 feat.)      & $108.0$ & $56.8$ & $89.8$ & $66.1$ \\
HistGB-Poisson (R-code feat.)      & $92.7$  & $53.4$ & $90.3$ & $66.1$ \\
HistGB-Poisson ($L_1$ Lasso, def.) & $108.5$ & $53.3$ & $83.6$ & $65.0$ \\
\textbf{HistGB-Poisson tuned ($L_1$, 20)} & $\mathbf{87.9}$ & $\mathbf{54.0}$ & $\mathbf{78.7}$ & $\mathbf{51.8}$ \\
\bottomrule
\end{tabular}
\end{table}


\begin{table}[!t]
\caption{Best result per feature selection method across CV strategies ($20$ features).}
\label{tab:ablate}
\centering
\footnotesize
\setlength{\tabcolsep}{3.5pt}
\renewcommand{\arraystretch}{1.05}
\begin{tabular}{llccc}
\toprule
\textbf{Selector} & \textbf{Best model} & \textbf{CV} & \textbf{RMSE} & \textbf{SMAPE} \\
\midrule
$L_1$ Lasso   & RandomForest    & 10-Fold           & $78.9$ & $\mathbf{52.7}$ \\
RF importance & HistGB\_Poisson & 10-Fold           & $\mathbf{78.5}$ & $53.9$ \\
F-regression  & RandomForest    & 10-Fold           & $83.9$ & $59.9$ \\
Mutual inf.   & RandomForest    & Rep $5{\times}10$ & $84.2$ & $58.5$ \\
\bottomrule
\end{tabular}
\end{table}

\paragraph{Finding 1: architecture matters, not only features.} On the identical R-code feature set, which isolates the model from any contribution of feature engineering, HistGB\_Poisson cuts holdout RMSE from $108.0$ to $92.7$ ($14\%$) and MAPE from $56.8\%$ to $53.4\%$.

\paragraph{Finding 2: feature selection is a precondition, not an option.} With all $92$ encoded inputs against $81$ training samples, the GLM attains a perfect training pseudo-$R^2$ of $1.000$ but produces test predictions averaging $254\times$ the true volumes, a textbook overfitting failure. Tree ensembles degrade more gracefully ($102.6$, $106.5$) but do not improve on Finding~1: irrelevant features dilute the signal.

\paragraph{Finding 3: $L_1$ Lasso is the strongest selector.} Lasso reaches RMSE $=79.8$ before any tuning, a $26\%$ reduction from the baseline achieved purely through a better subset. RF importance is a competitive second; mutual information and F-regression trail because they score features univariately. Under the GLM every selector still gives unusable error ($145$--$538$): selection alone cannot rescue the parametric model.

\paragraph{Finding 4: the sweet spot is $20$--$30$ features.} A sweep over $\{10,15,20,25,30\}$ features gives Lasso RMSE of $84.2$, $81.9$, $78.9$, $81.0$, and $79.8$ respectively. Below $15$, informative predictors are excluded; above $30$, irrelevant ones re-enter. We fix $20$ for interpretability.

\paragraph{Finding 5: tuning delivers the largest single gain.} Across $140$ tuning runs ($7$ models $\times$ $5$ feature counts $\times$ $4$ CV strategies, roughly $70{,}000$ fits), tuning lowers RMSE from $\sim 79$ to $73.7$ (Table~\ref{tab:tuning}). Two patterns recur: repeated $3{\times}10$ CV is selected by five of the seven tuned winners, and low learning rates ($0.01$) with moderate depth ($7$) and explicit $\ell_2$ regularization consistently dominate, all choices known to suppress overfitting on small data.

\begin{table}[!t]
\caption{Best configuration per model after tuning. Repeated $3{\times}10$ is chosen by five of the seven winners.}
\label{tab:tuning}
\centering
\footnotesize
\setlength{\tabcolsep}{3.5pt}
\renewcommand{\arraystretch}{1.05}
\begin{tabular}{llccccc}
\toprule
\textbf{Model} & \textbf{CV} & \textbf{\#f} & \textbf{RMSE} & \textbf{MAPE} & \textbf{SMAPE} \\
\midrule
HistGB\_Poisson & Rep $3{\times}10$ & $30$ & $\mathbf{73.7}$ & $66.4$ & $50.7$ \\
ExtraTrees      & Rep $3{\times}10$ & $20$ & $73.9$ & $74.3$ & $50.4$ \\
HistGB\_Squared & $10$-Fold        & $20$ & $74.5$ & $58.6$ & $\mathbf{46.8}$ \\
GradBoost\_Huber& Rep $3{\times}10$ & $15$ & $75.2$ & $69.9$ & $51.5$ \\
RandomForest    & Rep $3{\times}10$ & $20$ & $75.9$ & $89.2$ & $54.8$ \\
PoissonRegressor& $10$-Fold        & $15$ & $76.3$ & $65.0$ & $48.1$ \\
Bagging\_HistGB & Rep $3{\times}10$ & $30$ & $76.4$ & $65.1$ & $50.7$ \\
\bottomrule
\end{tabular}
\end{table}

\paragraph{Findings 6 and 7: two extensions fail.} Log-transforming the target harms HistGB\_Poisson badly ($73.7 \rightarrow 106.8$; HistGB\_Squared $94.4$, RF $81.8$, Extra Trees $77.2$): the Poisson loss already applies an internal log-link, so pre-transforming creates a double log-link that distorts the conditional mean. Practitioners using Poisson-loss models should not pre-log the target. Ridge-meta stacks of RF\,$+$\,HistGB\_Poisson\,$+$\,ExtraTrees and RF\,$+$\,HistGB\_Poisson\,$+$\,HistGB\_Squared give RMSE $76.1$ and $79.7$; the former reaches the best SMAPE we saw ($48.97$) but still trails the tuned single model. Tuning one strong learner beats combining several weaker ones here.

\paragraph{Selection and validation.} To avoid cherry-picking a metric, each of the $77$ candidates is scored under all four CV strategies by average rank on RMSE, MAPE, and SMAPE, retaining only those in the top decile under all four. Exactly one survives: HistGB\_Poisson with $L_1$ Lasso ($20$ features), tuned hyperparameters (learning rate $0.01$, $500$ boosting iterations, max depth $7$, minimum leaf size $10$, $\ell_2 = 1.0$), under repeated $3{\times}10$ CV, giving RMSE $=78.72$, MAPE $=51.79$, SMAPE $=49.93$. Holdout and CV agree on the winner; absolute values differ ($87.85$ vs.\ $78.72$) because the holdout rests on only $20$ test points. Since the target is heavy-tailed, we also repeated $3{\times}10$ CV with folds stratified on count quintiles: the ranking is unchanged and the winner's RMSE moves only to $81.05$.

\begin{figure}[!htbp]
\centering
\begin{tikzpicture}[every node/.style={font=\scriptsize}, xscale=0.032, yscale=0.185]
\draw[->, thick] (0,0) -- (92,0) node[right, font=\tiny] {$\Delta$RMSE};
\foreach \x in {0,20,40,60,80} {
  \draw (\x,-0.15) -- (\x,0.15) node[below=2pt, font=\tiny] {\x};
}
\foreach \v/\y/\l in {
  79.69/10/{stv\_mi},
  63.69/9/{sig\_int\_em},
  62.35/8/{dist\_CBD},
  54.27/7/{Retail\_Area\_em},
  49.06/6/{zch\_hm},
  41.55/5/{swlk\_len},
  29.21/4/{swlk\_len\_hm},
  27.79/3/{season\_fall},
  21.43/2/{Comm\_Area\_qm},
  18.53/1/{crossing\_unmark}
} {
  \fill[blue!55] (0,\y-0.34) rectangle (\v,\y+0.34);
  \node[anchor=east, font=\tiny] at (-1,\y) {\l};
  \node[anchor=west, font=\tiny] at (\v+1.0,\y) {\v};
}
\end{tikzpicture}
\caption{Top-$10$ permutation importances of the tuned HistGB\_Poisson model. Suffixes: \texttt{\_em} $=$ 1/8\,mi, \texttt{\_qm} $=$ 1/4\,mi, \texttt{\_hm} $=$ 1/2\,mi; unsuffixed $=$ intersection buffer.}
\Description{Horizontal bar chart of the ten most important features. Strava annual walking miles at the intersection ranks first at 79.69, followed by signalized intersections within an eighth of a mile at 63.69, distance to the central business district at 62.35, retail floor area at 54.27, zero-car households at 49.06, and sidewalk length at 41.55.}
\label{fig:imp}
\end{figure}
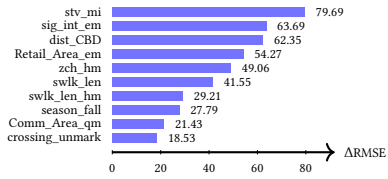

\paragraph{What the model uses.} Figure~\ref{fig:imp} reports permutation importance~\cite{strobl2007bias} of the $20$ Lasso-selected features: the increase in RMSE when a feature's values are shuffled. It describes how the boosting model uses a feature after selection, not the Lasso coefficient that selected it: \texttt{sig\_int\_em} has the largest Lasso coefficient ($+45.37$) but ranks second, behind \texttt{stv\_mi}. Crowdsourced walking activity, proximity to the central business district, signalization, retail floor area, zero-car households, and sidewalk length are precisely the correlates direct demand models have long identified~\cite{ewing2010meta}, supporting the conceptual validity of a model selected purely on predictive criteria.

\section{Conclusion}\label{sec:Conclusion}
We presented a systematic, reproducible machine learning pipeline for pedestrian volume estimation from GIS-derived built-environment features that takes the small-sample, high-dimensional regime seriously. Through approximately $70{,}000$ controlled model fits across $526$ configurations, we showed that a tuned Histogram-based Gradient Boosting model with Poisson loss, paired with $L_1$ Lasso feature selection ($20$ features) and Repeated $3{\times}10$ cross-validation, reduces RMSE by $27\%$ over the published Negative Binomial GLM baseline. We also showed that two common extensions, log-transformation and stacking, did not improve results in this setting. The selected predictors align with established pedestrian planning theory, supporting the conceptual validity of the model.

\begin{acks}
This work was funded by the Pacific Northwest Transportation Consortium (PacTrans), the Regional University Transportation Center (UTC) for Federal Region 10. We gratefully acknowledge its support.
\end{acks}

\bibliographystyle{ACM-Reference-Format}
\bibliography{references}

@article{boeing2017osmnx,
  author  = {Boeing, Geoff},
  title   = {{OSMnx}: New Methods for Acquiring, Constructing, Analyzing, and Visualizing Complex Street Networks},
  journal = {Computers, Environment and Urban Systems},
  volume  = {65},
  year    = {2017},
  pages   = {126--139},
  doi     = {10.1016/j.compenvurbsys.2017.05.004}
}

@article{breiman2001rf,
  author  = {Breiman, Leo},
  title   = {Random Forests},
  journal = {Machine Learning},
  volume  = {45},
  number  = {1},
  year    = {2001},
  pages   = {5--32},
  doi     = {10.1023/A:1010933404324}
}

@article{ewing2010meta,
  author  = {Ewing, Reid and Cervero, Robert},
  title   = {Travel and the Built Environment: A Meta-Analysis},
  journal = {Journal of the American Planning Association},
  volume  = {76},
  number  = {3},
  year    = {2010},
  pages   = {265--294},
  doi     = {10.1080/01944361003766766}
}

@article{friedman2001gbm,
  author  = {Friedman, Jerome H.},
  title   = {Greedy Function Approximation: A Gradient Boosting Machine},
  journal = {The Annals of Statistics},
  volume  = {29},
  number  = {5},
  year    = {2001},
  pages   = {1189--1232},
  doi     = {10.1214/aos/1013203451}
}

@article{geurts2006extremely,
  author  = {Geurts, Pierre and Ernst, Damien and Wehenkel, Louis},
  title   = {Extremely Randomized Trees},
  journal = {Machine Learning},
  volume  = {63},
  number  = {1},
  year    = {2006},
  pages   = {3--42},
  doi     = {10.1007/s10994-006-6226-1}
}

@article{pedregosa2011scikit,
  author  = {Pedregosa, Fabian and Varoquaux, Ga\"{e}l and Gramfort, Alexandre and Michel, Vincent and Thirion, Bertrand and Grisel, Olivier and Blondel, Mathieu and Prettenhofer, Peter and Weiss, Ron and Dubourg, Vincent and Vanderplas, Jake and Passos, Alexandre and Cournapeau, David and Brucher, Matthieu and Perrot, Matthieu and Duchesnay, \'{E}douard},
  title   = {Scikit-Learn: Machine Learning in {Python}},
  journal = {Journal of Machine Learning Research},
  volume  = {12},
  year    = {2011},
  pages   = {2825--2830}
}

@article{schneider2009pilot,
  author  = {Schneider, Robert J. and Arnold, Lindsay S. and Ragland, David R.},
  title   = {Pilot Model for Estimating Pedestrian Intersection Crossing Volumes},
  journal = {Transportation Research Record},
  volume  = {2140},
  number  = {1},
  year    = {2009},
  pages   = {13--26},
  doi     = {10.3141/2140-02}
}

@article{strobl2007bias,
  author  = {Strobl, Carolin and Boulesteix, Anne-Laure and Zeileis, Achim and Hothorn, Torsten},
  title   = {Bias in Random Forest Variable Importance Measures: Illustrations, Sources and a Solution},
  journal = {BMC Bioinformatics},
  volume  = {8},
  number  = {1},
  year    = {2007},
  pages   = {25},
  doi     = {10.1186/1471-2105-8-25}
}

@article{tibshirani1996lasso,
  author  = {Tibshirani, Robert},
  title   = {Regression Shrinkage and Selection via the Lasso},
  journal = {Journal of the Royal Statistical Society, Series B (Methodological)},
  volume  = {58},
  number  = {1},
  year    = {1996},
  pages   = {267--288}
}

@article{wang2022machine,
  author  = {Wang, Yuan and Hankey, Steve and Buehler, Ralph and Marshall, Julian D.},
  title   = {Predicting Bicycle and Pedestrian Traffic Volumes Using Mobile Sensor Data: A Machine Learning Approach},
  journal = {Transportation Research Part D: Transport and Environment},
  volume  = {110},
  year    = {2022},
  pages   = {103428},
  doi     = {10.1016/j.trd.2022.103428}
}

@inproceedings{golchin2026llm,
  author    = {Golchin, Bahareh and Rekabdar, Banafsheh and Justo, Danielle},
  title     = {{LLM}-Enhanced Reinforcement Learning for Time Series Anomaly Detection},
  booktitle = {2026 IEEE 20th International Conference on Semantic Computing (ICSC)},
  year      = {2026},
  address   = {Laguna Hills, CA, USA},
  publisher = {IEEE}
}

@inproceedings{golchin2025dynamic,
  author    = {Golchin, Bahareh and Rekabdar, Banafsheh},
  title     = {Dynamic Reward Scaling for Multivariate Time Series Anomaly Detection: A {VAE}-Enhanced Reinforcement Learning Approach},
  booktitle = {2025 IEEE International Conference on Cognitive Machine Intelligence (CogMI)},
  year      = {2025},
  address   = {Pittsburgh, PA, USA},
  publisher = {IEEE}
}

\end{document}